\documentclass[10pt,twocolumn,letterpaper]{article}

\usepackage{cvpr}              

\usepackage{multirow}
\usepackage{colortbl}
\newlength\savewidth\newcommand\shline{\noalign{\global\savewidth\arrayrulewidth\global\arrayrulewidth 1pt}\hline\noalign{\global\arrayrulewidth\savewidth}}

\definecolor{cvprblue}{rgb}{0.21,0.49,0.74}
\usepackage[pagebackref,breaklinks,colorlinks,allcolors=cvprblue]{hyperref}

\def\paperID{3248} 
\def\confName{CVPR}
\def\confYear{2025}

\title{MoVE-KD: Knowledge Distillation for VLMs with Mixture of Visual Encoders 
}

\author{
    Jiajun Cao\textsuperscript{\rm 1,2},
    Yuan Zhang\textsuperscript{\rm 1},
    Tao Huang\textsuperscript{\rm 3}, 
    Ming Lu\textsuperscript{\rm 1}, 
    Qizhe Zhang\textsuperscript{\rm 1},
    Ruichuan An\textsuperscript{\rm 1},\\
    Ningning Ma\textsuperscript{\rm 2},
    Shanghang Zhang\textsuperscript{\rm 1,$\ddagger$} \\
    \textsuperscript{\rm 1}State Key Laboratory of Multimedia Information Processing,\\School of Computer Science, Peking University
 \quad\\
    \textsuperscript{\rm 2}Autonomous Driving Development, NIO \quad
    \textsuperscript{\rm 3}Shanghai Jiao Tong University\\
    cao.jiajun888@gmail.com
}

\begin{document}
\maketitle
\begin{abstract}
Visual encoders are fundamental components in vision-language models (VLMs), each showcasing unique strengths derived from various pre-trained visual foundation models. To leverage the various capabilities of these encoders, recent studies incorporate multiple encoders within a single VLM, leading to a considerable increase in computational cost. In this paper, we present Mixture-of-Visual-Encoder Knowledge Distillation (MoVE-KD), a novel framework that distills the unique proficiencies of multiple vision encoders into a single, efficient encoder model. Specifically, to mitigate conflicts and retain the unique characteristics of each teacher encoder, we employ low-rank adaptation (LoRA) and mixture-of-experts (MoEs) to selectively activate specialized knowledge based on input features, enhancing both adaptability and efficiency. To regularize the KD process and enhance performance, we propose an attention-based distillation strategy that adaptively weighs the different encoders and emphasizes valuable visual tokens, reducing the burden of replicating comprehensive but distinct features from multiple teachers. Comprehensive experiments on popular VLMs, such as LLaVA and LLaVA-NeXT, validate the effectiveness of our method. Our code is available at:
\href{https://github.com/hey-cjj/MoVE-KD}{https://github.com/hey-cjj/MoVE-KD}.
\vspace{-7mm}

\end{abstract} 
\section{Introduction}
\label{sec:intro}

The rapid development of large vision-language models (VLMs) has significantly advanced artificial intelligence, particularly in tasks requiring integrated visual and linguistic understanding. At the core of these models, the vision encoder is essential for visual perception, forming the foundation for interpreting visual inputs and enabling the effective execution of vision-language tasks. Recent studies~\cite{shi2024eagle,tong2024cambrian, jia2024lift3dfoundationpolicylifting} highlight the distinct strengths of various vision encoders, such as CLIP~\cite{radford2021learning}, EVA~\cite{fang2024eva}, and ConvNeXt~\cite{liu2022convnet}, each excelling in specific vision-language applications. This diversity makes the optimization and integration of visual encoders a key area of research.


\begin{figure}[t]
    \centering
    \includegraphics[width=0.8\linewidth]{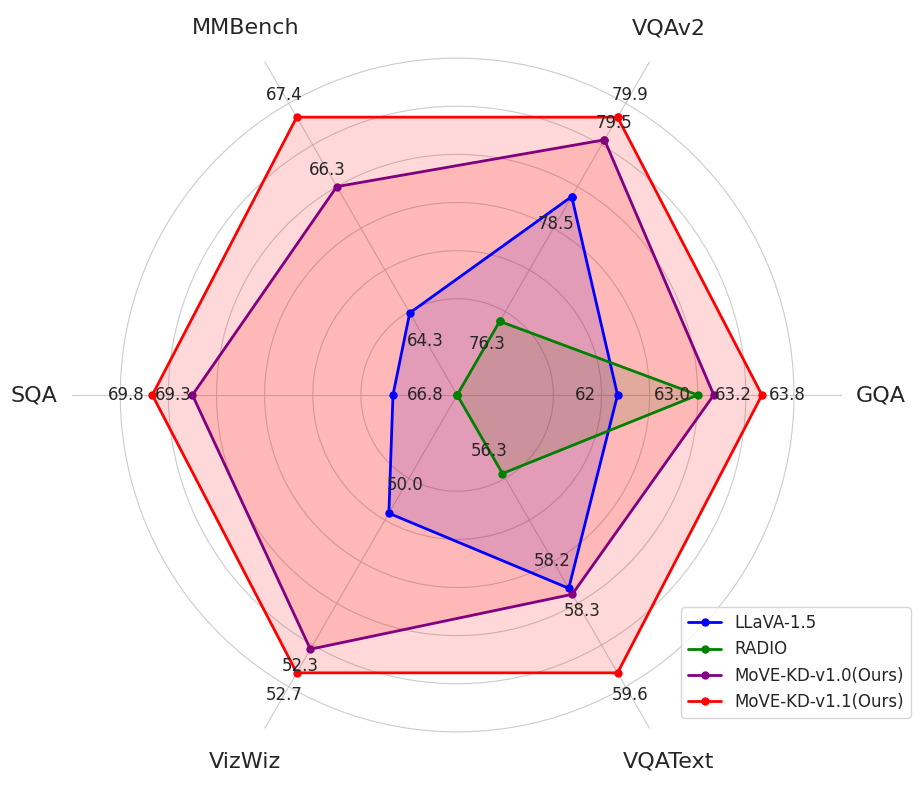}
    \caption{Comparison of LLaVA-1.5-7B~\cite{liu2023improvedllava} and RADIO~\cite{ranzinger2024radio} on a wide range of benchmarks, and MoVE-KD surpasses them.}
    \label{fig:radar}
    \vspace{-0.5cm}
\end{figure}

To harness the diverse proficiencies of various vision encoders, current methods~\cite{shi2024eagle,tong2024cambrian,li2024mini} often employ multiple encoders in a vision-language model via feature concatenation or attention mechanisms. However, compared to VLMs with a single vision encoder, using multiple encoders unavoidably increases computational costs and model complexity, diminishing efficiency and scalability. To address this, we explore a critical question in this paper: \textit{can we distill the unique proficiencies of various encoders into a single vision encoder, capturing their collective advantages while improving overall efficiency?}



To unify multiple encoders into one, knowledge distillation (KD)~\cite{hinton2015distilling} presents a promising approach, as it effectively transfers knowledge from a teacher model to a student model. However, classical KD methods primarily focus on one-to-one distillation, and the simultaneous distillation from multiple models, each with distinct pre-training datasets and objectives, remains relatively under-explored. Although AM-RADIO~\cite{ranzinger2024radio} proposes using multiple heads within a single model to replicate the predictions of various vision foundation models, its performance is constrained by conflicts arising from learning diverse and often competing characteristics within a shared backbone.


Our method fine-tunes a base model through knowledge distillation from multiple pre-trained visual foundation models, using a mixture-of-LoRA-experts (MoLE) framework. In this framework, the base model is adapted with multiple low-rank adaptation (LoRA) experts, which mitigate the catastrophic forgetting issue and can be selectively activated based on input characteristics. This design allows the model to dynamically harness strengths and specialized  insights of each teacher encoder, achieving a cohesive and efficient single-encoder structure.

\begin{figure}
  \centering
  \hspace{0mm}
  \begin{subfigure}{0.48\linewidth}
    \centering 
    \caption{Input image.}
    \includegraphics[width=1.0\linewidth]{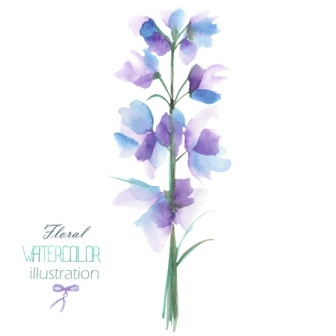}
    \label{fig:image}
    \vspace{0mm}
  \end{subfigure}
  \hfill
  \hspace{-2mm}
  \begin{subfigure}{0.48\linewidth}
    \centering 
    \caption{[\texttt{CLS}] attention map.}
    \includegraphics[width=1.0\linewidth]{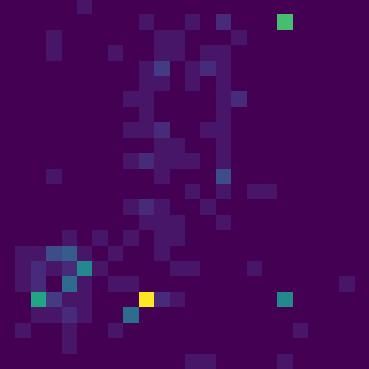}
    \label{fig:clip_atten}
    \vspace{0mm}
  \end{subfigure}
  \hfill
  \vspace{-4mm}
  \caption{Concentration of [\texttt{CLS}] attention. The left subfigure (a) is the input image, and the right subfigure (b) is the [\texttt{CLS}] attention visualization of the pre-trained CLIP, showing CLIP's focus on valuable regions of the image.}
  \label{fig:attn_dispersion}
  \vspace{-5mm}
\end{figure}

In addition to the student-side selective adaptation, it is crucial to identify and refine valuable features from the teacher models. Inspired by \cite{darcet2024vision}, which extends the input sequence of vision transformers with additional tokens to capture essential semantic attention, we propose using attention mechanisms to guide knowledge distillation from the teachers. Specifically, we use the [\texttt{CLS}] token to assess the importance of each visual token, applying a weighted distillation loss that prioritizes the valuable tokens from the teachers. Furthermore, we use the average importance of visual tokens as a weighting factor to balance the contributions of multiple teachers; in other words, teachers with higher average importance for a given sample are considered more influential to the accuracy. This selective distillation ensures that only valuable information from the teachers is absorbed by the student, effectively enhancing its ability to compress knowledge from multiple teachers into a single model.

Our final method, Mixture-of-Visual-Encoder KD (MoVE-KD), effectively integrates the strengths of multiple encoders while maintaining the efficiency of single-encoder models. Extensive experiments on popular VLMs and standard benchmarks demonstrate substantial improvements in both performance and efficiency over existing methods. Our contributions are as follows:

\begin{itemize}
\item We propose the MoVE-KD framework for multi-vision encoder fusion, marking the first approach to integrate different encoders for large vision-language models from a knowledge distillation perspective.

\item We introduce attention-guided KD regularization, which enhances the distillation of critical visual tokens and assigns adaptive weight to each teacher. Additionally, we incorporate Mixture-of-LoRA-Experts (MoLE) to prevent knowledge confusion.

\item Our framework has been applied to LLaVA and LLaVA-NeXT, achieving state-of-the-art performance across multiple benchmarks.
\end{itemize}

\section{Related Work}

\subsection{Vision-language models}
With the impressive success of large language models (LLMs) \cite{achiam2023gpt, yang2023baichuan, touvron2023llama, bai2023qwen, bi2024deepseek, young2024yi}, recent studies work on generative large vision-language models (VLMs) \cite{liu2023improvedllava, team2023gemini, Qwen-VL, chen2023internvl, li2024mini, ye2023mplug, lin2024draw, luo2025llm} to improve multimodal comprehension and generation through utilizing the strong generality of LLMs. 
Built upon the CLIP \cite{radford2021learning} image encoder which is somewhat aligned with the language modality, current VLMs typically utilize vast image-text pairs to connect the vision encoder and LLM, enabling LLM to receive and understand visual content. For instance, Flamingo \cite{alayrac2022flamingo} integrates visual features into LLM through gated attention. LLaVA \cite{liu2023improvedllava} directly connects the vision encoder and LLM with MLPs, showing proficiency in multi-modal dialogues. 
Besides, recent works boost the representation of the vision encoder \cite{li2024mini, shi2025we, zhang2024beyond} to further enhance the perception of VLMs. For example, Mini-Gemini \cite{li2024mini} employs an additional vision encoder for high-resolution refinement, and $S^{2}$ \cite{shi2025we} introduces multiple visual branches by scaling up the image scale.
However, the above methods are usually fed with higher-resolution image inputs or designed with extra modules, which require more computational resources. In our paper, we propose to improve the vision modality of VLMs through the adaptive supervision of a mixture of visual experts, where higher-quality training data is optional.

\begin{figure*}[!ht]
    \centering
    \includegraphics[width=0.85\linewidth]{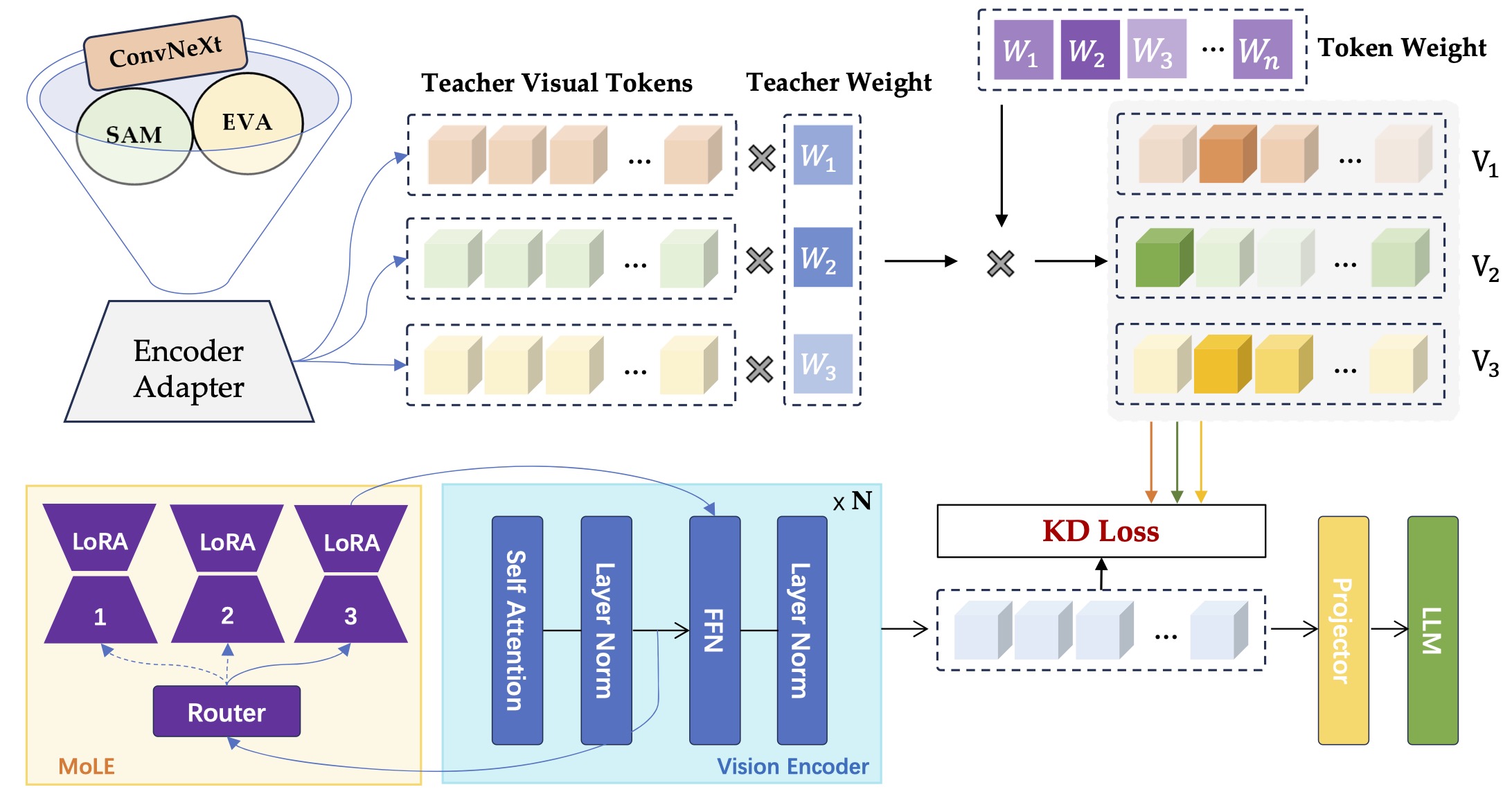}
    \caption{\textbf{The pipeline of MoVE-KD.} MoVE-KD projects teacher encoders' outputs using encoder adapters, assigns teacher weight and token weight based on CLIP's [\texttt{CLS}] attention. To mitigate knowledge conflicts, we incorporates MoLE structure in the student encoder. }
    \label{fig:pipeline}
    \vspace{-5mm}
\end{figure*}

\subsection{Knowledge distillation}
Knowledge distillation (KD) \cite{hinton2015distilling} is proposed to transfer dark knowledge in the teacher model to a student model, to boost the student without more parameters. Since the large vision-language models have become popular, how to enhance VLMs via KD is a notable research direction. 

\noindent \textbf{KD on vision encoder.} In vision-language models, the vision encoder is essential for extracting high-level features from visual signals, and provides perception and visual understanding ability. Before that, DINOv2 \cite{oquab2023dinov2} employs self-distillation to train their smaller variants from the larger teacher, while \cite{wei2022contrastive} distills their model from a CLIP teacher, which cannot inspire the potential of students due to the limited ability of the single teacher. In contrast, the AM-RADIO \cite{ranzinger2024radio} starts to utilize multiple vision experts to distill the vision encoder, and replace the original one in VLMs. However, these methods do not integrate into the VLMs framework, which are independent. Our paper is the first to combine distinct encoders for VLMs through knowledge distillation in a unified manner, which is beneficial to aligning vision modality and achieving better performance.

\noindent \textbf{KD with multiple teachers.} In traditional deep learning, \cite{you2017learning} starts to distill the student model under multiple teacher supervision with triplet loss. Most recently, OneS \cite{xue2022one} is the first to introduce multiple teachers into LLMs knowledge distillation, which gathers key knowledge from different pre-trained experts. Until now, to the best of our knowledge, we are the first to enhance VLMs by knowledge distillation with multiple vision teachers.
\section{Method}
\label{sec:method}
\subsection{Overview}
\label{sec:overview}
We propose MoVE-KD, a novel knowledge distillation method with multiple visual encoders for visual-language Models (VLM). The pipeline of MoVE-KD is shown in Fig. \ref{fig:pipeline}. Specifically, we first employ encoder adapters to project the outputs of multiple teacher encoders into a unified representation space. Based on the [\texttt{CLS}] attention from the pre-trained CLIP model, weights are dynamically assigned to both the teacher encoders and the visual tokens. Then, the KD loss is calculated based on the weighted sum of the teacher weights and token weights. To mitigate potential conflicts from learning multiple sources of knowledge, we incorporate a mixture-of-LoRA-experts (MoLE) structure within the student encoder. Our final training objective is to minimize the text loss and the KD loss.

\subsection{Learning from multiple encoders}
\label{sec:multi_encoder} 

\noindent \textbf{Encoder adapter.} Given a visual input, it is processed by different visual encoder teachers to obtain visual tokens. Due to the inconsistent representation spaces of visual encoders from different sources, these visual tokens cannot be directly aligned with the student visual token. Meanwhile, the commonly-used linear interpolation in KD methods is challenging to bridge the distinct tokens from different encoders to a unified and student-friendly token space, as discussed by LLaVA-1.5~\cite{liu2023improvedllava}.
Therefore, to match the dimensions and align the token spaces, we introduce encoder adapters for each teacher encoder. Each adapter, implemented as a two-layer MLP tailored to the output of its respective teacher, is independently utilized and optimized using the knowledge distillation loss.




\noindent \textbf{Mixture-of-LoRA-experts (MoLE).} After token alignment, the student encoder, initialized from the pre-trained CLIP visual encoder, is fine-tuned to learn teacher tokens. However, we find that directly fine-tuning the student encoder poses certain challenges:
Fine-tuning directly on the target dataset leads to issues like overfitting and catastrophic forgetting, which affect both model accuracy and generalizability (as detailed in Sec. \ref{sec:ablation}). Additionally, it is difficult to use the same shared weights to achieve a unified representation that retains all advantages while resolving conflicts across teacher tokens.

To address this challenge, we introduce the mixture-of-LoRA-experts (MoLE) architecture. The architecture contains two components: mixture-of-eperts (MoE) and low-rank adaptation (LoRA) expert. We first follow the typical design of MoE~\cite{jacobs1991adaptive} to selectively activate specific experts based on the inputs.
Formally, for each layer's FFN in the student encoder and input feature $x \in \mathbb{R}^{n \times d}$, the MoE output ${F}^{\star}(x)$ can be formulated as follows:
\begin{align}
    \begin{split}
        {F}^{\star}(x) &= {F}(x) + {E}_{i}(x)\\
        \text{with}\quad i &= \text{argmax}(\text{Softmax}(f(x))),
    \end{split}
\end{align}
where the router $f$ is a linear layer which learns weights to each expert, ${E}_{i}$ is the chosen $i$-th expert, and ${F}(x)$ is the original output of FFN.
This approach is particularly effective for knowledge distillation in multi-visual tasks by activating relevant expert, enhancing the model’s adaptability to diverse domains of visual knowledge.

Nevertheless, previous MoE works \cite{lin2024moe,shu2024llava,li2024uni} typically replicate FFN modules to serve as individual experts, which results in a substantial increase in parameter count and associated time costs. Therefore, our method uses parameter-efficient LoRA~\cite{hu2021lora} as the expert instead. By substituting one large parameter matrix with two low-rank matrices, LoRA significantly reduces the number of trainable parameters while maintaining performance. Besides, LoRA has also been demonstrated better generalizability and transferability \cite{xin2024beyond,asadi2024does}, which is particularly useful in our case where the encoder is fine-tuned with limited data.


Our MoLE facilitates the distillation process, allowing the model to better capture the strengths of each teacher while avoiding conflicts among their knowledge, with only a minor parameter overhead.

\renewcommand{\multirowsetup}{\centering}
\definecolor{mygray}{gray}{.92}
\definecolor{ForestGreen}{RGB}{34,139,34}
\newcommand{\fg}[1]{\mathbf{\mathcolor{ForestGreen}{#1}}}
\definecolor{Forestred}{RGB}{220,50,50}
\newcommand{\fr}[1]{\mathbf{\mathcolor{Forestred}{#1}}}
\definecolor{goldenrod}{RGB}{250, 250, 235}
\newcolumntype{a}{>{\columncolor{goldenrod}}c}
\begin{table*}[t]
    \centering
    \setlength{\tabcolsep}{2.8pt}
    \renewcommand{\arraystretch}{1.6}
    \footnotesize
	\centering
	
    \begin{tabular}{p{3.5cm} c | p{1cm}<{\centering} p{1cm}<{\centering} p{1cm}<{\centering} p{1cm}<{\centering} p{1cm}<{\centering} p{1cm}<{\centering} p{1cm}<{\centering} p{1cm}<{\centering}}
        \shline
        \textbf{Method} & \textbf{LLM} &\textbf{VQA}$^{\text{V2}}$ & \textbf{GQA} &  \textbf{VQA}$^{\text{Text}}$ & \textbf{VizWiz} & \textbf{POPE} & \textbf{SQA}& \textbf{MME} & \textbf{MMB} \\
        \hline
        \rowcolor{mygray}
        \multicolumn{10}{c}{\textit{1.7B Models}} \\
        LLaVA-1.5 \cite{liu2023improvedllava} & MobileLLaMA-1.4B \cite{chu2023mobilevlm}  & 71.5 & 55.4 & 42.6 & 28.6 & 84.3 & 56.0 & 1145.7 & 47.0 \\
        \rowcolor{goldenrod}
        + MoVE-KD-v1.0  & MobileLLaMA-1.4B \cite{chu2023mobilevlm}  & \underline{72.9} & \underline{56.6} & \underline{43.4} & \textbf{32.1} & \underline{84.8} & \underline{56.1} & \underline{1182.3} & \underline{47.4}\\
        \rowcolor{goldenrod}
        + MoVE-KD-v1.1  & MobileLLaMA-1.4B 
        \cite{chu2023mobilevlm}  & \textbf{73.8} & \textbf{57.7} & \textbf{44.3} & \underline{29.3} & \textbf{86.1} & \textbf{57.3} & \textbf{1188.4} & \textbf{48.8}\\
        \hline
        
        \rowcolor{mygray}
        \multicolumn{10}{c}{\textit{7B Models}} \\
        InstructBLIP \cite{liu2024visual} & Vicuna-7B \cite{zheng2023judging} & - & 49.2 & 50.1 & 34.5 & - & 60.5 & - & 36 \\
        Qwen-VL \cite{bai2023qwen} & Qwen-7B \cite{bai2023qwen} & 78.8 & 59.3 & 63.8 & 35.2 & - & 67.1 & 1487.5 & 38.2\\
        \hline
        LLaVA-1.5 \cite{liu2023improvedllava} & Vicuna-7B \cite{zheng2023judging} & 78.5 & 62.0 & 58.2 & 50.0 & 85.9 & 66.8 & 1510.7 & 64.3 \\
        + RADIO \cite{ranzinger2024radio} & Vicuna-7B \cite{zheng2023judging} & 76.3 & 63.0 & 56.3 & - & 86.2 & - & - & - \\
        \rowcolor{goldenrod}
        + MoVE-KD-v1.0  & Vicuna-7B \cite{zheng2023judging} & \underline{79.5} & \underline{63.2} & \underline{58.3} & \underline{52.3} & \textbf{86.9} & \underline{69.3} & \textbf{1524.5} & \underline{66.3} \\
        \rowcolor{goldenrod}
        + MoVE-KD-v1.1  & Vicuna-7B \cite{zheng2023judging} & \textbf{79.9} & \textbf{63.9} & \textbf{59.6} & \textbf{52.7} & \underline{86.3} & \textbf{69.8} & 1509.1 & \textbf{67.4} \\
        \hline
        LLaVA-NeXT \cite{liu2024llavanext} & Vicuna-7B \cite{zheng2023judging} & 81.8 & 64.2 & 64.9 & 57.6 & 86.5 & 70.1 & 1519.0 & 67.4 \\
        \rowcolor{goldenrod}
        + MoVE-KD-v1.0 \texttt{\scriptsize{(Ours)}} & Vicuna-7B \cite{zheng2023judging} & \textbf{82.3} & \textbf{64.5} & 63.7 & \textbf{58.0} & \textbf{86.7} & \textbf{70.7} & \textbf{1537.2} & \textbf{67.6} \\
        \hline

        \rowcolor{mygray}
        \multicolumn{10}{c}{\textit{13B Models}} \\
        InstructBLIP \cite{liu2024visual} & Vicuna-13B \cite{zheng2023judging} & - & 49.5 & 50.7 & 33.4 & 78.9 & 63.1 & 1212.8 & - \\
        \hline
        LLaVA-1.5 \cite{liu2023improvedllava} & Vicuna-13B \cite{zheng2023judging} & 80.0 & 63.3 & 61.3 & 53.6 & 85.9 & 71.6 & 1531.3 & 67.7 \\
        \rowcolor{goldenrod}
        + MoVE-KD-v1.0 & Vicuna-13B \cite{zheng2023judging} & \underline{80.6} & \textbf{64.2} & 59.7 & \underline{55.7} & 85.7 & \textbf{73.2} & \underline{1568.1} & \textbf{70.2} \\
        \rowcolor{goldenrod}
        + MoVE-KD-v1.1 & Vicuna-13B \cite{zheng2023judging} & \textbf{80.8} &  \underline{63.9} & \underline{61.1} & \textbf{57.5} & \textbf{86.3} & \underline{71.8} & \textbf{1568.3} & \underline{69.7} \\
        \hline
        LLaVA-NeXT \cite{liu2024llavanext} & Vicuna-13B \cite{zheng2023judging} & 82.8 & 65.4 & 67.1 & 60.5 & 86.2 & 73.6 & 1575.0 & 70 \\
        \rowcolor{goldenrod}
        + MoVE-KD-v1.0 \texttt{\scriptsize{(Ours)}} & Vicuna-13B \cite{zheng2023judging} & \textbf{83.1} & \textbf{65.7} & 65.8 & \textbf{60.9} & \textbf{86.8}  & \textbf{73.7} & \textbf{1579.3} & \textbf{70.6}  \\

        \shline
	\end{tabular}
     \vspace{-2mm}
     \caption{\textbf{Performance of MoVE-KD and other methods.} }
    \vspace{-2mm}
	\label{tab:main}
    \vspace{-3mm}
\end{table*}

\subsection{Attention-guided KD regularization}
\label{sec:cls_reg}

The key of distilling knowledge from multiple teachers into one model is to guide the student in which features should be focused on. Since different visual encoders have different understanding towards one image, and some representations are useless or redundant for the visual-language recognition, paying too much attention on those representations would weaken the learning of the real important and unique features. Therefore, a proper way is to find an appropriate constraint to regularize the distillation.


In such constraint, the student should be guided by distillation loss, which discriminates the valuable and redundant regions in the teacher tokens on both fine-grained token level and coarse-grained teacher level. Therefore, an ideal distillation loss on visual tokens can be formulated as
\begin{equation}
    \mathcal{L}_{kd} = \sum_{i=1}^{m} W^{(tea)}_{i} \sum_{j=1}^{n} (W^{(tok)}_{j}+\frac{1}{n}) \mathrm{MSE}(V^{(t)}_{i,j},V^{(s)}_{j}),
    \label{eq:kd_loss}
\end{equation}
where $m$ denotes the number of teacher encoders, $n$ is the sequence length of visual tokens, $V^{(t)}\in\mathbb{R}^{m\times n\times c}$ and $V^{(s)}\in\mathbb{R}^{n\times c}$ represent visual tokens of teacher and student, and $W^{(tok)}$ and $W^{(tea)}$ denote the token-level and teacher-level weight vectors.

Given the above motivation, we now elaborate on our method for deriving the weights $W^{(tok)}$ and $W^{(tea)}$. Basically, a good visual encoder should have a strong perception and focusing ability for the key information in the image. In this paper, instead of using the commonly-used learnable tokens \cite{huang2022masked} to capture the weights, we adopt a more efficient and generalizable way by using the [\texttt{CLS}] token in CLIP. As shown in Fig.~\ref{fig:attn_dispersion}, the cross-attention between the [\texttt{CLS}] token and other visual tokens in CLIP reveals the key regions in the image and shows less interest in repeated and unimportant information (see Sec. \ref{sec:discussion} for detailed discussion). This focusing characteristic, as well as the influential regions, would be beneficial for the student to learn from. Therefore, we design our KD regularization using the weights provided by the [\texttt{CLS}] attention of CLIP.

\begin{figure}[!t]
    \centering
    \includegraphics[width=0.8\linewidth]{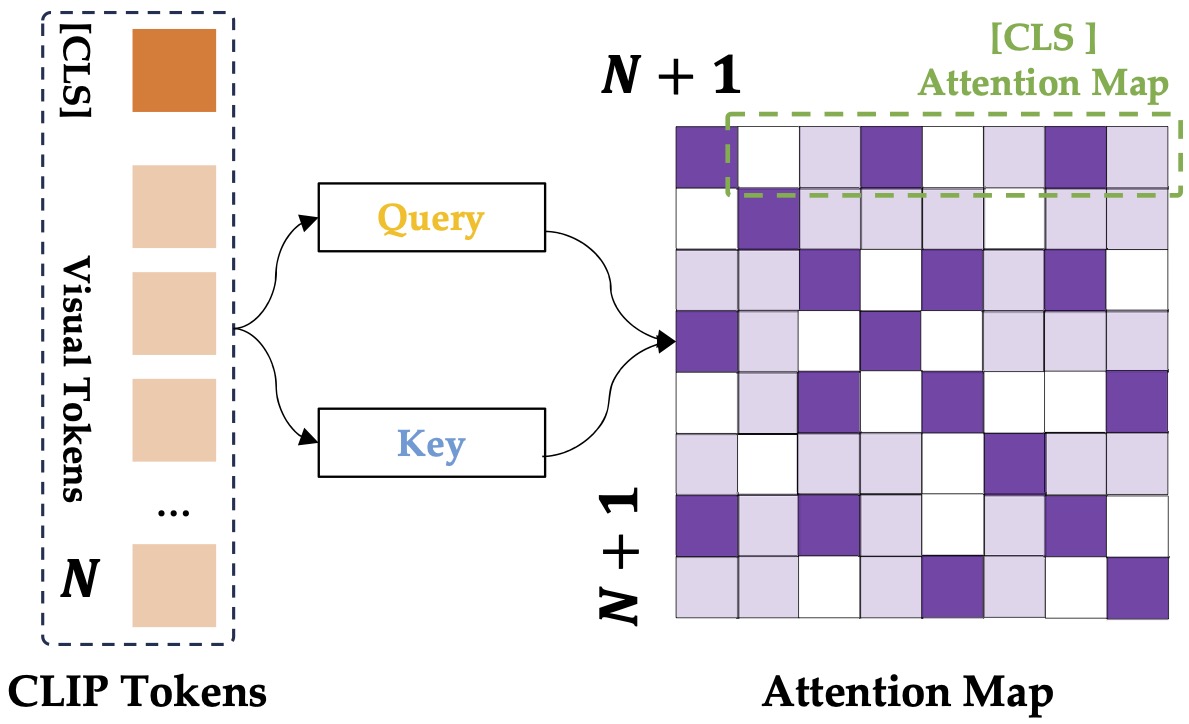}
    \caption{\textbf{The generation of token weight.} We employ the attention map of the [\texttt{CLS}] token to measure the contribution of vision tokens to knowledge distillation. }
    \label{fig:token_weight}
    \vspace{-5mm}
\end{figure}

\noindent \textbf{Token weight.} As mentioned above, we hope that the student encoder can focus on key visual tokens just like the pre-trained CLIP. Therefore, we calculate the [\texttt{CLS}] attention between the [\texttt{CLS}] token  $V^{(cls)} \in \mathbb{R}^{d}$ and other visual tokens $V^{(res)} \in \mathbb{R}^{n \times d}$ of CLIP, and use the normalization as the weight of each token, which is formulated as:
\begin{equation}
    W^{(tok)} = Softmax(\frac{(V^{(cls)} W^{(Q)})\cdot (V^{(res)} W^{(V)})^T}{\sqrt{d}})
    \label{eq:token_weight}
\end{equation}
where $W^{(tok)} \in [0,1]^{n}$ represents the token weight, while $W^{(Q)}$ and $W^{(V)}$ are the transformation matrices for queries and keys at this output layer of CLIP. Additionally, $d$ is a factor for stabilizing the values. 

\noindent \textbf{Teacher weight.} For the coarse teacher-level regularization, we take the Softmax of the mean value of the cross-attention between the [\texttt{CLS}] token ${V^{(cls)}}$ and the $i$-th teacher's tokens ${V^{(t)}_i}\in \mathbb{R}^{n \times d}$ as its weight. These weights of various teachers, indicate the responses of teachers to the specific images, and thus showing their contributions to the recognition, which is formulated as:
\begin{equation}
    W^{(tea)} = Softmax(mean(\frac{V^{(cls)}\cdot {V^{(t)}_i}^T}{\sqrt{d}}))
    \label{eq:teacher_weight}
\end{equation}
where $W^{(tea)} \in [0,1]^{m}$ is the teacher weight, and m is the number of teacher encoders. Note that our student is initialized by the pre-trained CLIP encoder, to prevent severe forgetting of its own knowledge, we involve CLIP as one teacher and set a relatively high fixed weight for it.

\subsection{Overall loss.}
\label{sec:loss}
The overall loss consists of two parts, namely the $\mathcal{L}_{text}$ and the $\mathcal{L}_{kd}$, representing the conventional log-likelihood loss in VLMs and distillation loss proposed in Eq. \ref{eq:kd_loss} of this paper, respectively.
As a result, the total loss is formulated as:
\begin{equation}
    \mathcal{L}_{total} = \mathcal{L}_{text} + \lambda_{kd} \cdot \mathcal{L}_{kd},
    \label{eq:total_loss}
\end{equation}
where $\lambda_{kd}$ is the weight of knowledge distillation.

\section{Experiments}

\renewcommand{\multirowsetup}{\centering}
\begin{table*}[t]
    \centering
    \setlength{\tabcolsep}{2.8pt}
    \renewcommand{\arraystretch}{1.6}
    \footnotesize
	\centering

    \begin{tabular}{p{3cm} | p{1cm}<{\centering} p{1cm}<{\centering} p{1cm}<{\centering} p{1cm}<{\centering} p{1cm}<{\centering} p{1cm}<{\centering} p{1cm}<{\centering} p{1cm}<{\centering} p{1cm}<{\centering}}
        \shline
        \textbf{Method}  &\textbf{VQA}$^{\text{V2}}$ & \textbf{GQA} &  \textbf{VQA}$^{\text{Text}}$ & \textbf{VizWiz} & \textbf{POPE} & \textbf{SQA}& \textbf{MME} & \textbf{MMB} & \ \textbf{Avg}\\
        \hline
        LLaVA-1.5 7b & 78.5 & 62.0 & 58.2 & 50.0 & 85.9 & 66.8 & 1510.7 & 64.3 & 66.5 \\
        ↑ KD (interpolation)  & 79.0 & 62.4 & 56.7 & 50.9 & 84.7 & 67.6 & 1507.6 & 62.9 & 66.3  \\
        ↑ Encoder adapter   & \underline{79.3} & 62.4 & 57.0 & 51.2 & 85.2 & 68.3 & 1517.8 & 63.8 & 66.7  \\
        ↑ MoLE & 79.1 & 62.8 & 57.3 & 51.9 & 86.4 & 68.7 & 1521.3 & 65.4 & 67.4  \\
        ↑ Token weight & \underline{79.3} & \underline{63.1} & \underline{57.7} & \textbf{52.5} & \underline{86.7} & \underline{68.9} & 1524.1 & \underline{66.0} & \underline{67.7}  \\
        ↑ Teacher weight & \textbf{79.5} & \textbf{63.2} & \textbf{58.3} & \underline{52.3} & \textbf{86.9} & \textbf{69.3} & \textbf{1524.5} & \textbf{66.3} & \textbf{68.0}  \\
       
        \hline
        \shline
	\end{tabular}
     \vspace{-2mm}
     \caption{\textbf{Ablation of MoVE-KD.} We add the designs in MoVE-KD one by one to explore the validity of each design. }
    \vspace{-2mm}
	\label{tab:ablation}
    \vspace{-3mm}
\end{table*}
In this section, we validate our method within various VLM architectures on comprehensive multimodal benchmarks to assess its effectiveness on image understanding tasks.

\subsection{Experimental settings}
\label{sec:exp_settings}
\noindent \textbf{Models.} 
We verify the proposed MoVE-KD on two popular VLM frameworks:  LLaVA \cite{liu2023improvedllava} and LLaVA-NeXT \cite{liu2024llavanext}. LLaVA-1.5 employs CLIP-pretrained \cite{radford2021learning} ViT-L as the visual tower. For resolution scaling, LLaVA-NeXT employs an adaptive image cropping strategy, encodes each image, and concatenates them in one single sequence. For LLaVA-1.5 and LLaVA-NeXT 7/13B, we follow the same training and inference setting as the original paper as it is available. LLaVA-1.5 1.7b is built on MobileLLaMA \cite{chu2023mobilevlm} 1.4b and is trained in the same way as LLaVA-1.5 7b.
For the teacher encoder, we retain CLIP \cite{wei2022contrastive} and additionally select EVA-02 \cite{fang2024eva} and ConvNeXt \cite{liu2022convnet}, which are mentioned in Eagle \cite{shi2024eagle} as the top performers on vision-language tasks.


\noindent \textbf{Dataset.} In the pre-training stage, we take LLaVA Visual Instruct Pretrain LCS-558K as the dataset. In the fine-tuning stage, the fine-tuning datasets of LLaVA-1.5 and LLaVA-NeXT are used respectively. For fairness, we do not introduce additional datasets for training.

\noindent \textbf{Benchmarks.}
To validate the effectiveness of our method, we conduct comprehensive experiments on eight widely adopted benchmarks including VQA$^{\text{V2}}$ (VQA V2) \cite{goyal2017making}, GQA \cite{hudson2019gqa}, VQA$^{\text{Text}}$ (TextVQA) \cite{singh2019towards}, VizWiz \cite{gurari2018vizwiz}, POPE \cite{li2023evaluating}, SQA \cite{lu2022learn}, MME \cite{fu2023mme}, MMBench (MMB) \cite{liu2023mmbench}. More details are included in the Appendix.

\noindent \textbf{Training details.} 
Our instruction-tuning procedure is consistent with LLaVA, consisting of two stages: pre-training and fine-tuning. In the pre-training stage, we tune the parameter weights of the student encoder’s MoLE, encoder adapters, and projection, while freezing all other parameter weights. In the fine-tuning stage, all parameter weights are updated except those of the teacher encoders. The training objective in both stages is to minimize the $\mathcal{L}_{total}$.

For hyperparameter settings, considering that the student encoder is initialized from a pre-trained CLIP, we assign a weight of 0.8 to the CLIP teacher encoder. The number of experts in MoLE is set to 3, and the rank $r$ of LoRA is set to 32. The weight for the distillation loss $\lambda_{kd}$ is set to 0.5. The other hyperparameters follow the settings of LLaVA.

The framework involves training on 16×A800 GPUs for standard machine configurations.

\subsection{Main results} 

To the best of our knowledge, MoVE-KD is the first KD method for visual encoders in VLMs, and there is no directly comparable baseline. When evaluating various benchmarks, we use the original model as the baseline. Additionally, RADIO \cite{ranzinger2024radio} is a known multi-encoder distillation method. However, it distills on DataComp-1B \cite{gadre2024datacomp} with 1.4 billion image-text pairs and then replaces CLIP in LLaVA-1.5 for training. Although the additional dataset is a bit unfair to our method, we still include it for reference.

Our results compared with the previous method are summarized in Tab. \ref{tab:main}, MoVE-KD achieves state-of-the-art (SOTA) performance on mainstream VLM frameworks like LLaVA-1.5 and LLaVA-NeXT. The RADIO \cite{ranzinger2024radio} exhibits obvious knowledge forgetting issues on VQA$^{\text{V2}}$ and VQA$^{\text{Text}}$, while our approach overcomes this problem and achieves comprehensive improvements on Viunca-7b \cite{zheng2023judging}. Besides, the LLaVA-1.5 equipped with our method even surpasses LLaVA-NeXT on some tasks, like MME and MMB, and it validates the effectiveness of MoVE-KD. 
Although we observe minimal degradation for MoVE-KD on VQA$^{\text{Text}}$, where a large number of questions are not related to vision. Our enhancement of vision may have a counterproductive effect on the model.

\vspace{-3mm}
       

\renewcommand{\multirowsetup}{\centering}
\begin{table}[!h]
    \centering
    \setlength{\tabcolsep}{1.8pt} 
    \renewcommand{\arraystretch}{1.4} 
    \footnotesize
	
    \vspace{2mm}
    
    \begin{tabular}{p{0.9cm}<{\centering} | p{0.9cm}<{\centering} p{0.8cm}<{\centering} p{1cm}<{\centering} p{0.9cm}<{\centering} p{0.8cm}<{\centering} p{0.7cm}<{\centering} p{0.7cm}<{\centering} p{0.6cm}<{\centering} }
        \shline
         \textbf{MoLE} &\textbf{VQA}$^{\text{V2}}$ & \textbf{GQA} &  \textbf{VQA}$^{\text{Text}}$ & \textbf{VizWiz} & \textbf{POPE} & \textbf{SQA} & \textbf{MMB} & \ \textbf{Avg}\\
        \hline
        w/o   & 76.7 & 60.0 & 54.8 & \textbf{51.9} & 81.5 & \textbf{64.1}  & \textbf{62.1} & 64.4 \\
        w/ & \textbf{77.4} & \textbf{60.3} & \textbf{55.6} & {51.3} & \textbf{82.1} & 63.7  & 61.5 & \textbf{64.5}  \\
       
        \hline
        \shline
	\end{tabular}
    \vspace{-2mm}
    \caption{\textbf{Impact of fine-tuning with MoLE introduction (without knowledge distillation).} }
    \label{tab:mole}
    \vspace{-3mm}
\end{table}

\renewcommand{\multirowsetup}{\centering}
\begin{table*}[t]
    \centering
    \setlength{\tabcolsep}{2.8pt}
    \renewcommand{\arraystretch}{1.6}
    \footnotesize
	\centering
	
    \vspace{-2mm}
	
    \begin{tabular}{p{2cm}<{\centering} | p{1cm}<{\centering} p{1cm}<{\centering} p{1cm}<{\centering} p{1cm}<{\centering} p{1cm}<{\centering} p{1cm}<{\centering} p{1cm}<{\centering} p{1cm}<{\centering} p{1cm}<{\centering}}
        \shline
         \textbf{CLIP weight} &\textbf{VQA}$^{\text{V2}}$ & \textbf{GQA} &  \textbf{VQA}$^{\text{Text}}$ & \textbf{VizWiz} & \textbf{POPE} & \textbf{SQA}& \textbf{MME} & \textbf{MMB} & \ \textbf{Avg}\\
        \hline
        0.6   & 78.8 & 62.6 & 57.3 & 51.7 & \underline{86.6} & 68.2 & 1518.4 & 64.5 & 67.1 \\
        0.7   & 79.1 & 62.5 & \underline{57.9} & \underline{52.1} & 86.3 & \textbf{69.5} & \underline{1521.7} & 65.2 & 67.5 \\
        0.8 & \underline{79.5} & \textbf{63.2} & \textbf{58.3} & \textbf{52.3} & \textbf{86.9} & 69.3 & \textbf{1524.5} & \textbf{66.3} & \textbf{68.0} \\
        0.9   & \textbf{80.1} & \underline{62.9} & 57.8 & 51.9 & \underline{86.6} & \underline{69.4} & 1519.5 & \underline{65.8} & \underline{67.8} \\
       
        \hline
        \shline
	\end{tabular}
     \vspace{-2mm}
     \caption{\textbf{Impact of the weight of CLIP teacher.} }
     \label{tab:clip_w}
     \vspace{-5mm}
\end{table*}

\subsection{Ablation study}
\label{sec:ablation}
We present the results of the ablation study for each design in our method in Tab. \ref{tab:ablation}. For simplicity, we use the LLaVA-1.5 7b model as the baseline and incrementally add each method design from MoVE-KD.

\noindent \textbf{Encoder adapter.}  Directly mapping the teacher's representation space to the student's through interpolation neglects feature continuity, which can easily lead to information loss. This approach can even result in average performance falling below the baseline. Through comparison, we demonstrated that utilizing learnable encoder adapters enables continuous feature mapping, ensuring effective knowledge transfer.

\noindent \textbf{Mixture-of-LoRA-experts.}  The introduction of MoLE allows the student encoder to dynamically select activated parameters based on the input, avoiding the knowledge confusion that arises during multi-teacher and multi-domain learning. This has led to significant performance improvements across various benchmarks, particularly in POPE and MMB. MoLE has mitigated the substantial performance degradation typically caused by knowledge distillation. Since the experts selected in MoLE are LoRAs, the parameters we introduce account for only 0.3\% of the total parameters. To eliminate the possibility that the performance improvement is solely due to the increase in parameters, we also conducted a control experiment where MoLE was introduced without knowledge distillation and the MoLE parameters were tuned during training. The results, as shown in Tab. \ref{tab:mole}, indicate that the introduction of MoLE does not lead to a performance improvement. On the contrary, some benchmarks show signs of degradation.

Additionally, we found that if MoLE is not introduced and instead the encoder is unfrozen for distillation, training often crashes (with the loss showing abnormal values). This situation typically requires setting an additional learning rate for the encoder. Compared to introducing MoLE, this approach is very inconvenient in practice.

\noindent \textbf{Attention-guided KD regularization.} Based on the regularization constraint guided by [\texttt{CLS}] attention, it plays a very good guiding role in the process of knowledge distillation, and further optimizes the model performance. As previously emphasized, the token weight is to better enable the student encoder to focus on key information. Constraints based on token weight can avoid dispersing the learning energy to background areas in the knowledge distillation process. What is said to be background will be further explained in the Sec. \ref{sec:background}. We visualize the [\texttt{CLS}] attention map of CLIP and the student in Fig. \ref{fig:attntion_visual}, which further illustrates the effectiveness of this constraint. It can be seen that under this constraint, the student does not lose its original focusing ability, and compared with CLIP, it further enhances the ability to grasp key information.

Since the student encoder is initialized by the pre-trained CLIP, as shown in Tab. \ref{tab:clip_w}, we find that it is necessary to give a higher weight to the CLIP teacher encoder in practice. What's more, due to the differences in the capabilities of different encoders, if the weights of the teachers are simply uniformized, it cannot allow each teacher to fully exert its strengths in its area of expertise. The teacher weight flexibly allocates the weights of each teacher according to the input, which indeed can improve the performance to a certain extent. However, the optimal method of teacher weight allocation needs further exploration.

\noindent \textbf{Unfreeze encoder.} Since unfreezing the encoder is necessary in our distillation process, there has been ongoing debate about whether unfreezing the encoder would enhance performance. To eliminate this factor's influence, we conducted experiments comparing LLaVA-1.5 with the encoder both unfrozen and frozen. As shown in Tab.~\ref{tab:unfreeze_encoder}, unfreezing the encoder clearly leads to performance degradation, demonstrating that our performance improvements are not related to unfreezing the encoder.

\noindent \textbf{Better performance with more teachers.
} To further demonstrate the scalability of MoVE-KD and the impact of additional teachers on the performance, we employ the SAM-L \cite{kirillov2023segment} as a new teacher alongside the original ones. We denote the two versions as MoVE-KD-v1.0 and MoVE-KD-v1.1. As shown in Table \ref{tab:main}, MoVE-KD-v1.1 further improves performance and outperforms MoVE-KD-v1.0 on VQA$^{\text{Text}}$, with MoVE-KD-v1.1 7B even surpassing LLaVA-1.5 13B on some benchmarks. 
The performance is further improved as the number of teachers increases, demonstrating a strong scalability with MoVE-KD. 

\subsection{Visualization} 
\begin{figure*}[!h]
    \centering
    \includegraphics[width=0.9\linewidth]{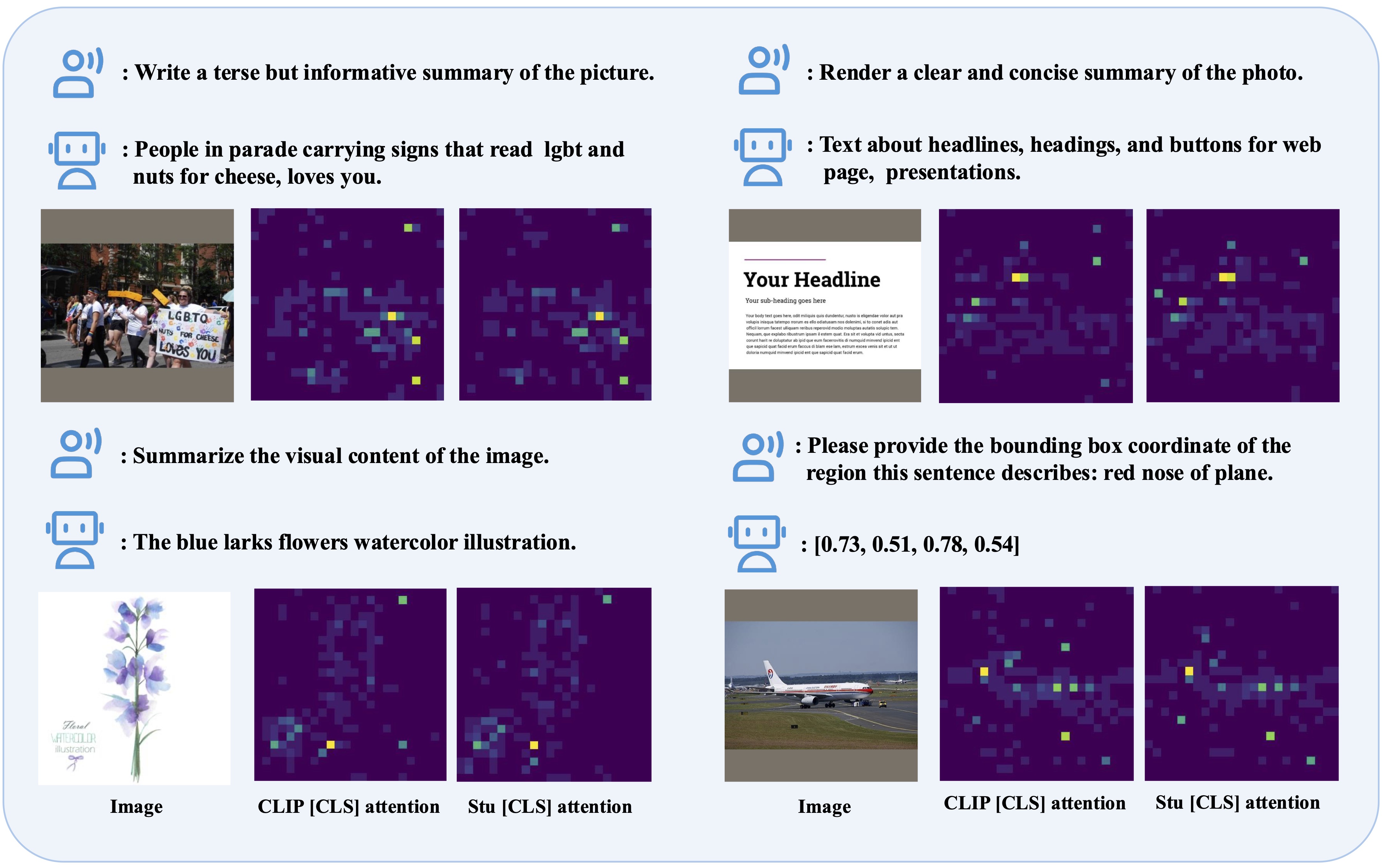}
    \caption{\textbf{The visualization of CLIP [\texttt{CLS}] attention and student [\texttt{CLS}] attention.} }
    \label{fig:attntion_visual}
    \vspace{-5mm}
\end{figure*}

In Fig. \ref{fig:attntion_visual}, we show the visualization results of the [\texttt{CLS}] attention of both the pre-trained CLIP and the distilled student model. In the upper left figure, the student encoder focuses more attention on the crowd and banner, reducing attention to the background above compared to CLIP. In the upper right figure, the student encoder maintains additional attention on the ``Headline'' and ignores blank areas. In the lower left figure, the student encoder disregards the blank area in the lower right, concentrating attention on the flowers and text. In the lower right figure, unlike CLIP, the student encoder clearly outlines the plane and the airport runway, without excessively focusing on the sky as CLIP does.

       

\renewcommand{\multirowsetup}{\centering}
\begin{table}[!h]
    \centering
    \setlength{\tabcolsep}{1.8pt} 
    \renewcommand{\arraystretch}{1.4} 
    \footnotesize
	\vspace{0mm}
    \begin{tabular}{p{1.3cm}<{\centering} | p{1cm}<{\centering} p{1cm}<{\centering} p{1cm}<{\centering} p{1cm}<{\centering} p{1cm}<{\centering} p{1cm}<{\centering}  }
        \shline
         \textbf{Encoder} &\textbf{VQA}$^{\text{V2}}$ & \textbf{GQA} &  \textbf{VQA}$^{\text{Text}}$ & \textbf{VizWiz} & \textbf{POPE} & \textbf{SQA} \\
        \hline
        Unfreeze  & 76.7 & 60.0 & 54.8 & \textbf{51.9} & 81.5 & 64.1  \\
        Freeze & \textbf{78.5} & \textbf{62.0} & \textbf{58.2} & {50.0} & \textbf{85.9} & \textbf{66.8}   \\
       
        \hline
        \shline
	\end{tabular}
    \vspace{-2mm}
    \caption{\textbf{Freeze vs unfreeze vision encoder.} }
    \vspace{-2mm}
	\label{tab:unfreeze_encoder}
    \vspace{-2mm}
\end{table}

\section{Discussion}
\label{sec:discussion}
In this section, we first explore the definitions of foreground and background in knowledge distillation of visual encoders, and then discuss the practical implications of visual tokens in the background with high [\texttt{CLS}] attention.
\subsection{Foreground and background definitions}
\label{sec:background}
In vision-language tasks, the image's foreground refers to the visual tokens most relevant to the text. SparseVLM~\cite{zhang2024sparsevlm} reduces the number of visual tokens by pruning low-similarity background visual tokens, calculated based on text-token and visual-token similarity. We also experimented with using this similarity as the weight for visual tokens during distillation to focus attention on the foreground. However, we found that this approach led to performance degradation. Although the text provides clear foreground information, the definition of foreground and background varies with different questions. For instance, in Fig. \ref{fig:attntion_visual}, the buildings behind a crowd may also become the foreground depending on the question context. Therefore, the definition of foreground and background in distillation should be fixed. When observing, humans tend to focus on dynamic, complex, semantically rich elements, while being less sensitive to repetitive items (\eg, sky, water, and grasslands). Hence, we propose using human-like foreground and background definitions as a regularization constraint for visual encoder knowledge distillation, where dynamic and semantic regions are treated as foreground, and regions with excessive repetitive content are treated as background. As shown in Fig. 
\ref{fig:clip_atten}, we find that CLIP’s [\texttt{CLS}] attetion indeed provides a relatively accurate foreground and background definition, aligning well with human visual perception.

\subsection{The high [\texttt{CLS}] attention tokens in background}
In Fig. \ref{fig:attntion_visual}, we observe some tokens in the background with significantly high [\texttt{CLS}] attention, which are referred to as ``artifacts" in the literature \cite{darcet2024vision}. The study suggests that exposing these artifacts can lead to more interpretable attention maps and improve performance in dense prediction tasks. However, as previously discussed, the background is not unimportant; rather, it can be summarized with fewer tokens to capture repetitive information, similar to peripheral vision in the human eye. We believe these so-called artifacts essentially carry rich global information, which is crucial for VLMs to generate accurate responses. CLIP condenses background regions with repetitive information using few tokens with relatively high attention. Therefore, we retain the [\texttt{CLS}] attention weight of these tokens.

\section{Conclusion}

In this paper, we introduce a novel framework called Mixture-of-Visual-Encoder Knowledge Distillation (MoVE-KD) aimed at fusing the unique proficiencies of multiple visual encoders into a single efficient encoder model, marking the first approach to integrate different encoders for large vision-language models from a knowledge distillation perspective. Through the use of low-rank adaptation (LoRA) and mixture-of-experts (MoEs) to selectively activate specialized knowledge based on input features, we successfully mitigate conflicts and preserve the distinctive characteristics of each teacher encoder. Our attention-based distillation strategy further enhances performance by adaptively weighing different visual encoders and emphasizing valuable visual tokens. Comprehensive experiments on popular VLMs like LLaVA and LLaVA-NeXT have validated the efficacy of our approach. Furthermore, as the scale of large language models increases, we notice diminishing marginal returns from knowledge distillation, suggesting that the performance bottleneck of large vision-language models may lie in the projector that bridges the visual encoder and the large language model (LLM). This indicates that further research into developing more efficient methods to seamlessly and losslessly project visual and text tokens into a unified representation space should be a key focus for future advancements in VLM research.

\section*{Acknowledgement}

This work was supported by the National Natural Science Foundation of China (62476011).

{
    \small
    \bibliographystyle{ieeenat_fullname}
    \bibliography{main}
}


\end{document}